%% file: iclr2020_ai4ah_handhygiene.tex
\documentclass{article} 
\usepackage{iclr2020_conference,times}

\input{math_commands.tex}

\usepackage{hyperref}
\usepackage{url}
\usepackage{graphicx}

\title{Fully Automated Hand Hygiene Monitoring\\in Operating Room using 3D Convolutional Neural Network}

\author{Minjee Kim \\
Department of Biomedical Engineering \\
University of Ulsan College of Medicine\\
Asan Medical Center, Seoul, South Korea \\
\texttt{minjeekim00@gmail.com} \\
\And
Joonmyeong Choi \\
Department of Medicine \\
University of Ulsan College of Medicine\\
Asan Medical Center, Seoul, South Korea \\
\texttt{jm5901@gmail.com} \\
\AND
Namkug Kim \\
Department of Convergence Medicine \\ 
University of Ulsan College of Medicine \\
Asan Medical Center, Seoul, South Korea \\
\texttt{namkugkim@gmail.com}
}

\iclrfinalcopy 
\begin{document}

\maketitle

\begin{abstract}
Hand hygiene is one of the most significant factors in preventing hospital acquired infections (HAI) which often be transmitted by medical staffs in contact with patients in the operating room (OR). Hand hygiene monitoring could be important to investigate and reduce the outbreak of infections within the OR. However, an effective monitoring tool for hand hygiene compliance is difficult to develop due to the visual complexity of the OR scene. Recent progress in video understanding with convolutional neural net (CNN) has increased the application of recognition and detection of human actions. Leveraging this progress, we proposed a fully automated hand hygiene monitoring tool of the alcohol-based hand rubbing action of anesthesiologists on OR video using spatio-temporal features with 3D CNN. First, the region of interest (ROI) of anesthesiologists' upper body were detected and cropped. A temporal smoothing filter was applied to the ROIs. Then, the ROIs were given to a 3D CNN and classified into two classes: rubbing hands or other actions. We observed that a transfer learning from Kinetics-400 is beneficial and the optical flow stream was not helpful in our dataset. The final accuracy, precision, recall and F1 score in testing is 0.76, 0.85, 0.65 and 0.74, respectively.
\end{abstract}

\section{Introduction}
Hand hygiene has been considered as the most crucial action to prevent infections such as hospital acquired infections (HAI). Many studies support that appropriate hand hygiene shows significant reduction of infections and enhances a patient safety in hospitals. As a part of a campaign, healthcare organizations have suggested the guidelines of hand hygiene. Despite of those efforts, the rates of hand hygiene compliance remain poor in most countries \citep{Berrera11, Loftus2019HandHI}.

The lack of compliance with hand hygiene can also be observed among medical staffs in the operating room (OR). Anesthesiologists, in particular, are highly involved in numerous activities that can affect HAI such as intubating a patient, accessing intravenous catheters, injecting of anesthetic drugs and airway management. However, the potential risk of HAI from anesthesiologist's hand disinfection has been underestimated.

A recent study shows that repeated monitoring and positive feedback on hand hygiene increased compliance with the recommended guidelines. An effective monitoring tool, however, is still difficult to find in hospital. The standard method of monitoring hand hygiene is conducted through a direct observation by trained observers within a specific monitoring period. This method is highly labor-intensive and subject to biases with the nature of sampling and the Hawthorne effect \citep{Srigley2014QuantificationOT}, the effect that people modify their behaviors by perception of being observed.

We proposed a fully automated hand hygiene monitoring tool using two-stream inflated 3D CNN, I3D \citep{Carreira17}. We leveraged a transfer learning from the network learned from Kinetics-400 \citep{Kay2017TheKH}, a large video dataset for human action classification, containing 400 categories with approximately 300,000 video clips. Also, we investigated the effect of optical flow stream for small-scale hand actions. Our study can further be expanded to analyze workflow patterns of hand hygiene providing a holistic understanding of hand hygiene protocols of ORs.

\section{Related Work}

A few previous studies have been conducted to develop an effective hand hygiene monitoring tool in hospital. For ORs, \cite{Bellaard12} have implemented a mobile-based tool that can instantly record the hand hygiene compliance during direct observations. This tool helps to collect quantitative data but could not overcome the sampling bias of the direct observation. In ICUs, \cite{Zhang2017ApplyingML} utilized the activation data of smart hand sanitizers to predict hand hygiene compliance. In outpatient settings, \cite{Geilleit18} proposed a real-time notification system by building a virtual line with an infrared sensor. These methods generally have a high accuracy of use of hand sanitizers yet require multiple sensors which are not suitable for ORs. Vision-based automatic monitoring tools with top-down view cameras are implemented by \cite{Yeung2016VisionBasedHH} and \cite{Haque17}, which limits the monitoring to the location of hand sanitizer dispensers. 

A recent progress of deep learning architectures in video domain enables to extract spatio-temporal features of human actions in video-level \citep{Tran2014LearningSF, Hara2017CanS3}. The semantic features of human actions can be utilized to monitor and assess hand hygiene compliance in ORs.

\section{Operating Room Dataset}
\subsection{Data Acquisition}
\begin{figure}[h]
    \centering
    \graphicspath{ {./figures/} }
    \includegraphics[scale=0.5]{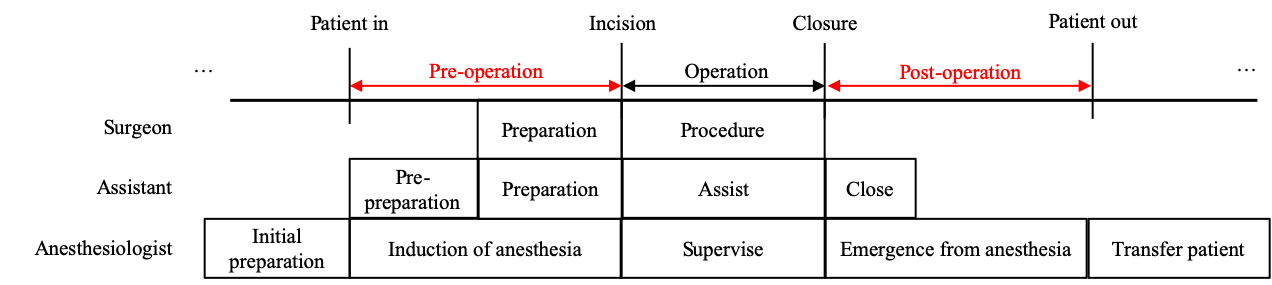}
    \caption{An example of work process of the surgical team (medical staffs) in the OR. Induction phase is the transition from an awake state to an anesthetized state and emergence phase is from the sleep state to full consciousness. We included only pre- and post-operative procedures in the dataset.
}
    \label{fig:0}
\end{figure}

The Institutional review board for human investigations at the cooperative hospital approved the retrospective study with a waiver of informed consent. The videos were collected for four months from a single OR of the hospital. The video recording was performed throughout multiple surgeries consecutively. We then selected pre- and post-operative scenes only as shown in Figure \ref{fig:0}, in which anesthesiologists are intensively involved. Faces of medical staffs and patients were blurred in all the images used for training. The videos were recorded at 15 fps in 640x480 resolution.

\subsection{Synthetic Data}
Synthetic data is often combined with real-world data to address the lack of available dataset. In our study, the deficiency was possibly derived from poor hand hygiene compliance, annotation missing and confined field of view of the camera. We additionally recorded an hour-long video under a simulated situation close to a real surgical settings. Accompanied with an anesthesiologist, eight people who are not medical staffs repeatedly rubbed their hands with the hand sanitizers installed in the same OR. The synthetic data was included in a training dataset only and a detail number of video clips is stated in \ref{data_stats}.

\subsection{Data Annotation}
To build upon a dataset that holds meaningful actions related to hand hygiene protocols, we annotated actions including 4 classes: rubbing hands, intubating, wearing and removing gloves. Among untrimmed videos, we selectively chose video clips containing those actions. Since our aim is to detect an alcohol-based hand rubbing action of anesthesiologists in the OR, we labeled rubbing hands as 1 and the other actions as 0.

\subsection{Medical Staff Detection}

\begin{figure}[h]
    \centering
    \graphicspath{ {./figures/} }
    \includegraphics[scale=0.5]{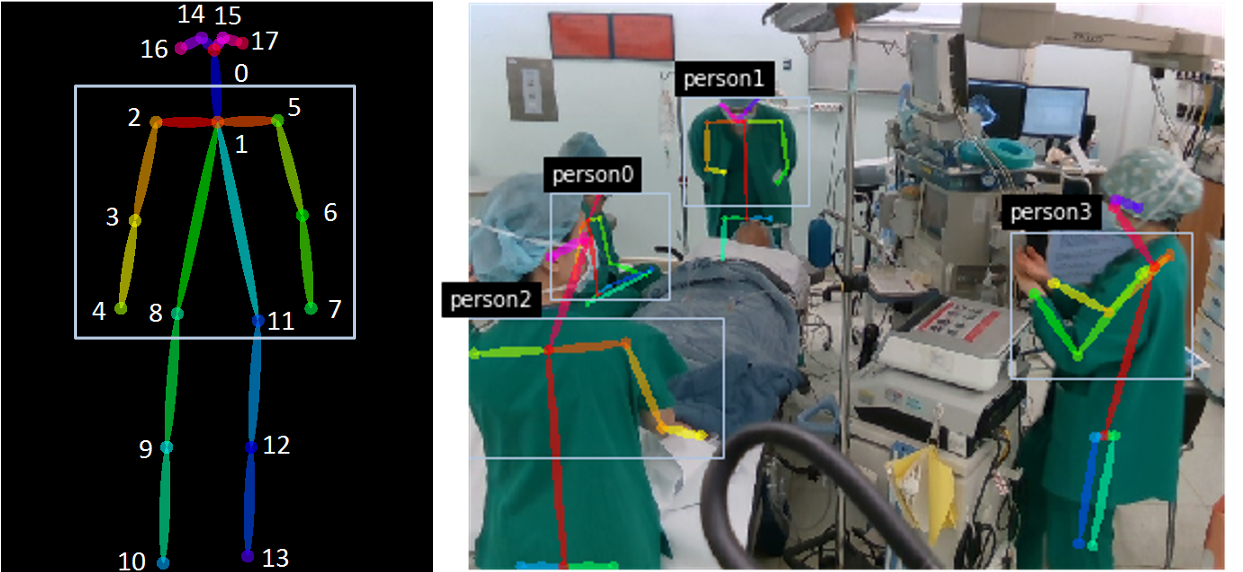}
    \caption{(Left) COCO-18 key-point format for human pose skeleton. (Right) The rendered skeleton outputs of pose estimation. The upper body ROIs were calculated with six key-points of two shoulders, elbows and wrists.}
    \label{fig:3}
\end{figure}

We utilized a real-time pose estimation library, OpenPose \citep{Cao_2017_CVPR}, which automatically extracts 2d human body key-points of multi-person on single images. We chose a COCO-18 format with x, y coordinates of 18 body parts from eye to ankle as standard. Since hand rubbing is mostly focused on upper body, we calculated upper body ROIs using six key-points of two shoulders, elbows and wrists and added some margins as shown in \ref{fig:3}. The coordinates were further used as data augmentation during training as detailed in \ref{data_aug}.

\subsection{Action Linking and Temporal Smoothing}
 In order to link the ROIs to the same person along with a video clip, we calculated the intersection over union (IoU) for every detected person in the next image and assigned it with the person with the highest score. False assignments were manually corrected. Additionally, we applied a simple moving average (SMA) filter to reduce temporal shakes of the ROIs. The equation of SMA is:
\begin{center}
   $\displaystyle y[n] = \sum_{k=0}^{L-1} x[n-k] $\\
\end{center}
where $L$ is the duration of discrete time of input vector $x$ and $y$ is the output vector of the average of the previous $L$ vectors. We set $L$ to be 4 which derived empirically.

\subsection{Data Statistics}
\label{data_stats}
A total of 45 untrimmed videos were acquired, of which 19 videos were confirmed to contain the target action, rubbing hands. The remaining videos were ignored. Each untrimmed video is eight hours in length and has one to eight occurrences of hand rubbing action. The total video clips are composed of 176 clips, of which 97 clips were labeled as rubbing hands, of which 79 clips were labeled as other actions. Each clip has the number of image frames with a range from 16 to 152. We split all video clips into 10:1:1 for training, validation and testing, respectively. All datasets keep the ratio of each class and do not share the video clips recorded in the same date.

\begin{table}[h!]
\centering
\begin{center}
\begin{tabular}{c||cccc|cc}
\hline
\multicolumn{1}{c||}{\bf }  &\multicolumn{4}{c}{\bf Real-world Data} &\multicolumn{2}{|c}{\bf Synthetic Data} \\
\multicolumn{1}{c||}{} &\multicolumn{2}{c}{Hand Rubbing} &\multicolumn{2}{c}{Other Actions} &\multicolumn{2}{|c}{Hand Rubbing} \\ 
\multicolumn{1}{c||}{Augmentation} &\multicolumn{1}{c}{\bf w/o} &\multicolumn{1}{c}{\bf w/} &\multicolumn{1}{c}{\bf w/o} &\multicolumn{1}{c}{\bf w/} &\multicolumn{1}{|c}{\bf w/o} &\multicolumn{1}{c}{\bf w/} \\\hline 
Training         &10(360)&214(3421)&67(4070)&1055(16880)&71(1799)&736(11776)\\
Validation       &8(231)&111(1776)&5(510)&148(2368)&0(0)&0(0)\\
Testing          &8(232)&112(1792)&7(402)&102(1632)&-&-\\
\hline
\end{tabular}
\caption{Description of the number of the video clips (the image frames) filmed in a real world and synthetic data. Other actions include intubating, wearing, removing gloves, etc. The number of data with augmentation was calculated using only temporal method, not spatial method stated in \ref{data_aug}}.
\end{center}
\end{table}

\section{Methods}
\subsection{Data Augmentation}
\label{data_aug}
 To overcome the dearth of dataset, we used data augmentation techniques spatially and temporally. Spatially, a real-time multi-scale cropping around the upper body region was applied. We randomly selected areas that are 1 to 1.75 times of the size of the original coordinates containing the upper body region. In addition, we applied a random horizontal flipping with a probability of 0.5 and brightness jittering with an extent from -0.1 to 0.1. For testing, we used the upper body cropping with the original ROI size. All of the input was scaled to 224 pixel size and center-cropped. Temporally, for rubbing hands class, we picked all possible starting frames in a single clip enough to have a consecutive 16 frames. For other actions class, we set a step size of 4 in frames between each clip. None of the temporal techniques were applied in testing.

\subsection{Hand Rubbing Action Classification}

We chose I3D model, an Inception-v1 architecture \citep{Szegedy2014GoingDW} that inflates 2D into 3D convolutions. Two I3D networks for RGB and optical flow inputs were separately and jointly trained on our dataset to explore the effect of optical flow modality. The optical flow was computed with a TV-L1 algorithm implemented in OpenCV library. The I3D was pretrained on a large video dataset for human action classification, Kinetics-400, with 400 categories including 9 hand-related human actions: air drumming, applauding, clapping, cutting nails, doing nails, drumming fingers, finger snapping, pumping fist and washing hands. The pretrained I3D was then fine-tuned with an additional 1x1x1 convolution after the last convolutional network and a sigmoid output was used to account for the binary classification of hand rubbing action. All the layers before and including the average pooling layer were frozen when training. We chose the number of input frames to be 16 to give to this architecture. 

\section{Experimental Results}
\subsection{Hand Rubbing Action Classification}

We set a training experiment on RGB stream I3D model as baseline and compared with optical flow model which trained separately. Then, we evaluated and compared three models including RGB, optical flow (Flow) and the pooling model (RGB+Flow) using I3D (Table \ref{table:2}). All methods were evaluated per video clip in testing.

\begin{table}[h!]
\centering
\begin{center}
\begin{tabular}{ccccc}
\hline
\multicolumn{1}{c}{Model} 
 & Accuracy & Precision & Recall & F1 score  \\\hline 
I3D (RGB)       &\bf 0.76	&0.85&	\bf0.65&	\bf0.74 \\
I3D (Flow)      & 0.62	&\bf1.00&	0.22&	0.36\\
I3D (RGB+Flow)  & 0.61	&0.94&	0.28&	0.43\\
\hline
\end{tabular}
\caption{Ablation studies for classification of hand rubbing using RGB, optical flow (Flow) and the joint model (RGB+Flow) using I3D.}
\label{table:2}
\end{center}
\end{table}

Table \ref{table:2} shows that RGB model which trained separately results in the best performance among three models, outperforming in accuracy, recall and f1 score. The result suggests that optical flow model is not helpful when trained separately and jointly when using our dataset. By pooling the model with optical flow stream, we observed 0.15 decrease in accuracy.

\begin{figure}[h]
    \centering
    \graphicspath{ {./figures/} }
    \includegraphics[scale=0.8]{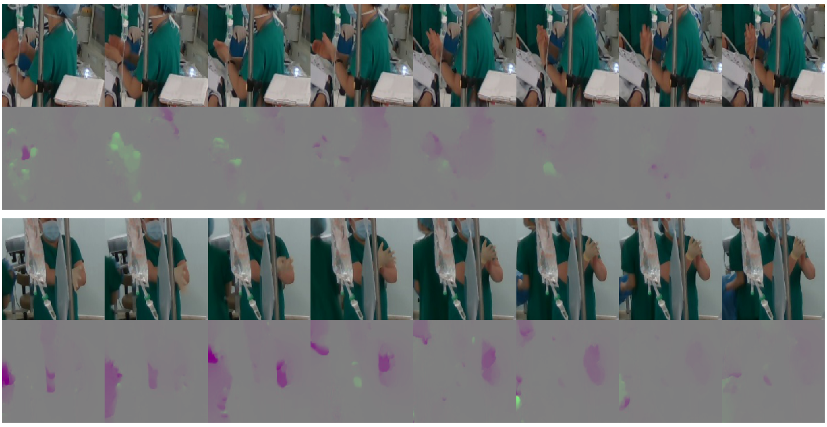}
    \caption{Small-scale actions of hand rubbing and other hand-related actions (glove wearing) in RGB and optical flow modality. (Top) Input images of hand rubbing action. (Bottom) False positive example in RGB model in testing. Input images of glove wearing action.
    }
    \label{fig:6}
\end{figure}

\section{Discussions \& Limitations}
We demonstrate that hand hygiene actions of medical staffs in OR can automatically be detected and monitored using spatio-temporal features with 3D CNN. It is encouraging that the challenges of small video datasets can be solved with the significant benefits of a pre-trained network trained on a large video dataset. The contribution of optical flow modality is much less useful in a small-scale hand action dataset than a large-scale action dataset.

We now analyze the results of optical flow model. Optical flow model alone was not able to properly predict the actions, scoring a prefect precision of 1.0 and a low recall of 0.22, which means the model returned none of the target class, rubbing hands. Previous studies have demonstrated that two-stream architectures have superior performance on many human action classification benchmark datasets \citep{Carreira17}, but the pooled model on our dataset resulted in 0.15, 0.37 and 0.31 decrease in accuracy, recall and f1 score, respectively. This gap of the impact of optical flow model reflects the different characteristics among the datasets. In benchmark datasets such as UCF-101 \citep{Soomro2012UCF101AD}, HMDB-51 \citep{Kuehne2011HMDBAL}, and Kinetics-400, the majority of the video clips are either with a single person performing large-scale actions like running and golf-swing or with a distinctive object or background like biking and playing cello. On the other hand, our dataset consists of the video clips with multi person performing very subtle motions of hands (Figure \ref{fig:6}.)

There are several limitations to this study. Abundant videos collected from different ORs should be trained to generalize our model in unseen surgical setup. The category of other actions needs to be expanded to analyze other hand hygiene-related activities such as touching patients or surgical instrument in ORs. Moreover, occlusion of hand location should be considered.

\section{Conclusions}
We proposed a fully automated hand hygiene monitoring tool in ORs using 3D CNN. With a combination of pose estimation and cropping upper body regions, we were able to detect the medical staff in ORs. We additionally applied a temporal smoothing filter after linking upper body ROIs along with a video clip. Synthetic data, a transfer learning and data augmentation using body key-points were utilized to overcome the small size dataset. The experimental results have demonstrated that I3D network with RGB stream outperforms optical flow and the joint model.
\bibliography{iclr2020_ai4ah_handhygiene}
\bibliographystyle{iclr2020_conference}

\end{document}

%% file: math_commands.tex
\usepackage{amsmath,amsfonts,bm}

\def\eqref#1{equation~\ref{#1}}

\def\1{\bm{1}}

\DeclareMathAlphabet{\mathsfit}{\encodingdefault}{\sfdefault}{m}{sl}
\SetMathAlphabet{\mathsfit}{bold}{\encodingdefault}{\sfdefault}{bx}{n}

